\documentclass[nohyperref]{article}

\def\isaccepted{}

\usepackage{microtype}
\usepackage{graphicx}
\usepackage{subfigure}
\usepackage{booktabs} 

\usepackage{hyperref}

\usepackage{icml2022}

\usepackage{amsmath}
\usepackage{amssymb}
\usepackage{mathtools}
\usepackage{amsthm}
\usepackage{array}
\newcolumntype{H}{>{\setbox0=\hbox\bgroup}c<{\egroup}@{}}

\usepackage[capitalize,noabbrev]{cleveref}

\usepackage{graphicx}

\theoremstyle{plain}

\theoremstyle{definition}

\theoremstyle{remark}

\usepackage[textsize=tiny]{todonotes}

\icmltitlerunning{Bayesian PPI - Page~\thepage~of~\pageref{page:end}}

\begin{document}

\twocolumn[
\icmltitle{Bayesian Prediction-Powered Inference}



\icmlsetsymbol{equal}{*}

\begin{icmlauthorlist}
\icmlauthor{R. Alex Hofer}{goog}
\icmlauthor{Joshua Maynez}{goog}
\icmlauthor{Bhuwan Dhingra}{goog}
\icmlauthor{Adam Fisch}{goog}
\icmlauthor{Amir Globerson}{googr}
\icmlauthor{William W. Cohen}{goog}
\end{icmlauthorlist}

\icmlaffiliation{goog}{Google DeepMind}
\icmlaffiliation{googr}{Google Research}

\icmlcorrespondingauthor{William Cohen}{wcohen@google.com}

\icmlkeywords{question answering, NLP}

\vskip 0.3in
]




\begin{abstract}
Prediction-powered inference (PPI) is a method that improves statistical estimates based on limited human-labeled data.  Specifically, PPI methods provide tighter confidence intervals by combining small amounts of human-labeled data with larger amounts of data labeled by a reasonably accurate, but potentially biased, automatic system.
We propose a framework for PPI based on
Bayesian inference that allows researchers to
develop new task-appropriate PPI methods easily.  
Exploiting the ease with which we can design new metrics,
we propose improved PPI methods for several important
cases, such as autoraters that give discrete responses (e.g., prompted LLM ``judges'') and autoraters with scores that have a non-linear relationship to human scores.   
\end{abstract}

\section{Introduction}


\subsection{Motivation for prediction powered inference}

Large foundation models often make it possible to build applications with very little training data.  However, even zero- and few-shot models do require \emph{evaluation} data.  In particular, to iteratively improve a method during development, or to confidently report an improvement, one needs to confidently and quantitatively assess performance of the method.  This is hard to do with evaluation sets of a few hundred examples or fewer.  

One often-proposed approach to avoiding the evaluation bottleneck is to use secondary LLM-based system to judge the output of the primary one: for instance, if the primary task is developing an LLM-based question-answering (QA) system, one could use a second LLM-based system that rates question/answer pairs as acceptable or not \cite{bohnet2023attributed, kamalloo-etal-2023-evaluating, bulian-etal-2022-tomayto}.  However, as others have noted \cite{doi:10.1126/science.adi6000}, automated raters may be biased relative to the human raters they are intended to model.

\begin{figure}[t]
\begin{center}
    \includegraphics[width=3.2in]{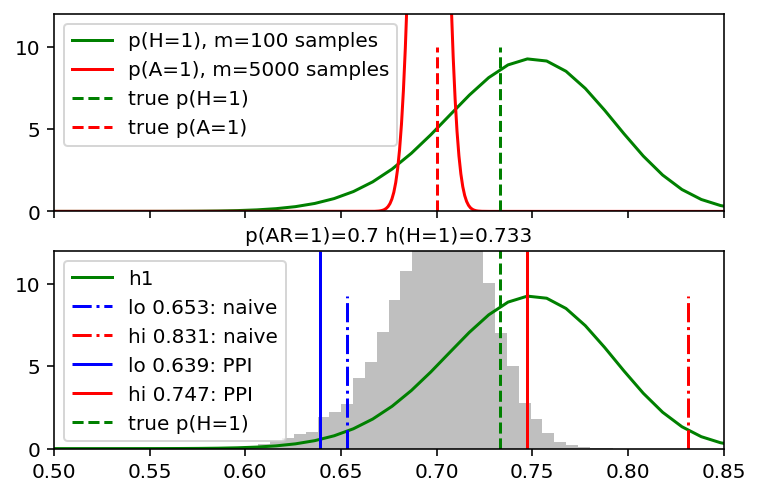}
\end{center}
\caption{Top: Estimating accuracy $P(H=1)$ with 100 human-labeled examples (green) or 5000 autorater-labeled examples (red). Dotted vertical lines are the true accuracies $P(H=1)$ and $P(A=1)$ (for this synthetic data). Bottom: The dot-dashed blue/red lines are a 95\% confidence interval computed with classical methods from 100 human-labeled examples. The grey histogram and solid blue/red lines are a 95\% confidence interval using PPI, which combines the autorater and human predictions (see text).}
\label{fig:h100-vs-ar-5000}
\end{figure}

To illustrate this, Figure~\ref{fig:h100-vs-ar-5000} (top) compares ratings from humans and an autorater for a hypothetical QA system.  The true\footnote{Known precisely here because we are using synthetic data.} rate of acceptable responses from the QA system as measured by humans is $P(H=1)=0.733$ (dashed green vertical line), but using only 100 human-labeled examples gives large variance: the uncertainty of the estimate (green curve) means that the true $P(H=1)$ may be plausibly be as low as 0.65 or as large as 0.82 (with 95\% confidence, it is between these values).  With an autorater, more examples can be labeled, leading to less uncertainty in the estimate (the red curve).  Unfortunately, the autorater gives very small probability to the true accuracy of 73.3\%, because it is \emph{biased}, with a true mean (red dashed line) of $0.7$.

Our choices seem to be between the high-variance unbiased estimate from a small human sample, and a lower-variance but biased autorater-based estimate.  However, there are also statistically valid ways of \emph{combining} the auto-rater and human data.  
Following \cite{doi:10.1126/science.adi6000} we will call such methods \emph{prediction-powered inference (PPI)} methods.  The grey histogram in the graph of Figure~\ref{fig:h100-vs-ar-5000} is a PPI method which gives a confidence interval which is much smaller than the ``naive'' classical interval, and which correctly contains the true value of $P(H=1)$.

\subsection{Bayesian inference for PPI}

In past work in PPI \cite{doi:10.1126/science.adi6000,angelopoulos2023ppi} the computation of a \emph{estimand}---e.g., a statistic to be computed, such as a population mean---is reduced to solving a convex optimization problem---e.g., finding $\mu$ that minimizes $\sum_{i} (y_i - \mu)^2$ for a sample $y_1, \ldots, y_n$.  Analytic techniques are then used to find bounds on the minimized value, typically by establishing asymptotic normality.  This approach is very natural for certain tasks---such as bounding coefficients of linear or logistic regressions---but less natural for others.

\newcommand{\twid}[1]{\tilde{#1}}
\newcommand{\expect}[1]{\mathbb{E}[{#1}]}

Here we propose an alternative approach to PPI, based on Bayesian inference. An advantage of our approach is that it allows researchers to easily design a new PPI estimand that makes task-appropriate use of autoraters, since much of the analysis can be replaced by general-purpose numerical methods that compute confidence intervals over the designed statistics.

As an illustration, consider estimating a mean value of some human label $y$ that evaluates an instance $x$ (e.g., an ``acceptability'' rating for a question/answer pair). We assume a small sample $S_n=(x_1,y_1),\ldots,(x_n,y_n)$ of points $x_i$ where $y_i$ is known, and a larger sample $\twid{S}_N=x_1,\ldots,x_N$ of $N \gg n$ unlabeled points $x_j$.  We also assume an \emph{autorater} or judge model $f(x)$, which predicts $y$, and our goal is to predict the expected value $\expect{y}$ as precisely as possible given the data  $D=(S_n, \twid{S}_N)$.

The natural estimand for $\expect{y}$ is the mean of the $y$'s in $S_n$.  An alternative is the following \emph{proxy estimand:}
\begin{eqnarray} \label{eq:diff-intro}
 g(S_n, \twid{S}_N) & = & \mu_1 + \mu_2, \mbox{~~~where} \\
 \mu_1 & = & \frac{1}{N}\sum_{j=1}^N f(x_j) \nonumber \\
 \mu_2 & = &  \frac{1}{n}\sum_{i=1}^n y_i - f(x_i)  \nonumber  
\end{eqnarray}  
Here $\mu_1$ is the mean autorater score over the unlabeled data,
and $\mu_2$ is the mean difference between the autorater and the human label over the labeled data $S_n$, called a \emph{rectifier} in \cite{doi:10.1126/science.adi6000}.  Because the rectifier corrects the bias of the autorater,  $\expect{g(S_n,\twid{S}_N)}$ is the same as $\expect{y}$, but it may have substantially lower variance if $N$ is large and $f(x)$ is accurate: 
$\mu_1$ will have low variance when $N$ is large, and the rectifier will  have low variance when $y_i - f(x_i)$ is generally small.

The statistic of Eq~\ref{eq:diff-intro}, sometimes called the \emph{difference estimate}, is well-studied and its variance can be computed analytically \cite{doi:10.1126/science.adi6000, difference-est-survey17}, leading to a classical confidence interval for $\expect{y}$.  We propose instead a Bayesian analysis, where we treat $\mu_1, \mu_2$ as random variables that depend on the data. Since $\mu_1$ and $\mu_2$ are independent given the data $D$\footnote{Note $\mu_1$ and $\mu_2$ are computed from  different subsets of $D$.}
we can marginalize them out as follows to compute the expectation of $g$ over the random variables:
\begin{equation} \label{eq:bayes-diff-est}
\expect{g(S_n, \twid{S_N})} 
  = \int (\mu_1 + \mu_2) p(\mu_1|D) p(\mu_2|D) d\mu_1 d\mu_2
\end{equation}
Using \emph{Monte Carlo integration} (see Sec~\ref{sec:mci} here or Sec.~24.2 in \cite{wasserman2013all}) it is possible to compute such intervals, and also to compute upper and lower bounds $\ell,u$ such that
$\Pr(\ell \leq \expect{g} \leq u) \leq c$ for a given confidence level $c$.  These bounds are called \emph{credible intervals} in Bayesian statistics (see Sec~\ref{sec:credible}).

This computation is very efficient, and can be used for a wide variety of potential proxy estimands, thus allowing easy implementation of new PPI-like methods.

\subsection{Contributions}

To summarize our contributions, we introduce a Bayesian framework for PPI tasks, specifically advocating for Monte Carlo integration as the fundamental inference process.  This leads immediately to a framework in which one can readily design autorater-powered proxy estimands for different tasks, and compute confidence intervals over these designed estimates.

Concretely, we propose and evaluate
\begin{itemize}
\item Bayesian variants of the difference estimate (called simply PPI in \cite{doi:10.1126/science.adi6000}) and its extension using powertuning (called PPI++ in \cite{angelopoulos2023ppi});
\item a Bayesian extension of the difference estimate called \emph{stratified estimates}, which improve experimentally over prior methods on several experimental tasks, and which are especially powerful when $n$ is of moderate size (a few hundred) and autorater scores have a  non-linear relationship to human labels;
\item combinations of stratified estimates with the ``power tuning'' approach of \cite{angelopoulos2023ppi}; 
\item a family of novel estimates we call \emph{chain rule estimates},
which improve substantially over difference estimates when autoraters give discrete, uncalibrated responses, which is common when autoraters are based on prompted LLMs.
\end{itemize}
We show that these new methods offer practically important improvements on a wide range of tasks, including judging outputs of summarization systems;  evaluating attributed question-answering systems; evaluating open-book QA systems; and conducting side-by-side tests on QA systems.

We also discuss and experimentally analyze the PPI-based credible intervals with classical confidence intervals, showing that the intervals produced by Bayesian difference estimates are virtually identical to their classical counterparts, and that the methods perform experimentally with respect to classical frequentist goals.


\section{Background}


\subsection{Classical confidence intervals and credible intervals} \label{sec:credible}

\subsubsection{Definitions}

The following material is provided for completeness and to establish notation, but can be found in many textbooks, such as \cite{wasserman2013all}.

Statistical \emph{confidence intervals} measure uncertainty of an estimate of an unknown parameter $\theta$ from a finite sample $X$---for example, $\theta$ might be the unknown value of $p(H=1)$ above.  The most familiar procedure for computing a confidence interval is to pick an appropriate \emph{probability distribution function (pdf)}, written here $f(\theta)$: recall that a pdf for $\theta$ has the property that $p(a \leq \theta \leq b) = \int_a^b f(\theta) d\theta$.  A \emph{confidence interval} for \emph{confidence level} $c$ is a pair of values $\ell$ and $u$ such that 
\begin{equation} \label{eq:bounds}
    \int_{\ell}^u f(\theta) d\theta = c
\end{equation}
An \emph{equal-tailed interval} discards the same amount of probability mass below $\ell$ and $u$, i.e., letting $\alpha = 1-c$, an equal-tailed interval is one where
\[ \int_{-\infty}^\ell f(\theta) d\theta  = \int_u^{\infty} f(\theta) d\theta  = \alpha/2
\]
We consider only equal-tailed intervals in this paper.

In a Bayesian setting, we typically think of $f(\theta)$ as the posterior of a random variable $\theta$ given data $D$, $f(\theta)=p(\theta|D)$, and the interval $\ell, u$ for level $c$ means that $p(\ell \leq \theta \leq u) > c$, where the probability is taken over the possible values of the unknown parameter $\theta$.
When this interpretation is being used, $\ell, u$ is called a \emph{credible interval}.  Since the true prior $p(\theta)$ is typically not known, it is generally necessary to use an \emph{improper} weak prior for $\theta$.

The more common classical (frequentist) interpretation of confidence intervals is more complicated. We can still say that $p(\ell \leq \theta \leq u) > c$, but the probability is taken over \emph{possible samples} $S$, and $\theta$ is assumed to be fixed (but unknown).

\subsubsection{Common examples}
\emph{Intervals for means.} One familiar case of this is when $\theta$ is the mean of an unknown normal distribution.  Given a sample $S=y_1,\ldots,y_n$, with sample mean 
$\overline{y}=\frac{1}{n} \sum_i y_i$
and sample variance $\hat{\sigma}^2 = \frac{1}{n} \sum_i (x_i - \overline{y})^2$, 
a confidence interval can be found using Eq~\ref{eq:bounds} by making $f(\theta)$ a Gaussian distribution $\mathcal{N}(\mu, \sigma)$, for $\mu=\overline{y}$ and 
\( \sigma = \sqrt{\hat{\sigma}^2 / n} \).  In this case there is a simple closed-form solution for the classical confidence interval, obtained by snipping off the tails of $f(\theta)$, which for a 95\% confidence interval gives
$\ell = \mu - 1.96\sigma$ and $u = \mu + 1.96\sigma$.  
 
It turns out that this $f(\theta)$ can also be interpreted as a posterior for $\theta$. If this is done the Bayesian credible intervals will be the same as the classical ones, and we follow this practice here.

\emph{Intervals for proportions.} Another common case is when the data is generated by a Bernoulli distribution, i.e., for each $x_i$ we have $p(x_i=1) = \theta$ and $p(x_i=1) = 1 - \theta$, for $0 \leq \theta \leq 1$.  It is common to use the same procedure as above---although the usual notation in describing it is to write $k$ for the number of times $x_i=1$, $\hat{p}=\hat{\mu}=k/n$, and  
$\hat{\sigma}^2=\hat{p}(1 - \hat{p})$.  This is called the Wald method.

The Wald method is an approximation and is \emph{conservative}---i.e., the intervals can be too large.  The most noticeable errors are when $\ell < 0$, which can happen when  $n$ is small and $\theta*$ is close to zero, or when $u> 1$, which can happen when $n$ is small and $\theta*$ is close to one.  Unfortunately this is an important case for the methods described below, so in the experiments below we use as the ``classical confidence interval'' the Clopper-Pearson test \cite{nelson2018five}.

Bayesian credible interval computations for proportions usually use a Beta distribution\footnote{Recall that Beta$(\alpha, \beta)$ is defined over $[0,1]$ and is closely related to the binomial distribution.}  as a prior, often the Jeffrey's prior of Beta$(\alpha=\frac{1}{2}, \beta=\frac{1}{2})$, and we follow this practice in the experiments of this paper.

A very useful extension to the Bernoulli distribution is the multinomial distribution, where the random variable $x_i$ can take on $K$ possible integer values 1, \ldots, $K$, so $\theta$ is a vector on a $K$-dimensional simplex.  The analog of the Beta distribution here is called the Dirichlet distribution, for which we use the prior $\alpha_1=\ldots\alpha_K = \frac{1}{K}$.


\subsection{Monte Carlo integration} \label{sec:mci}

\begin{figure*}
\begin{center}
    \includegraphics[width=\textwidth]{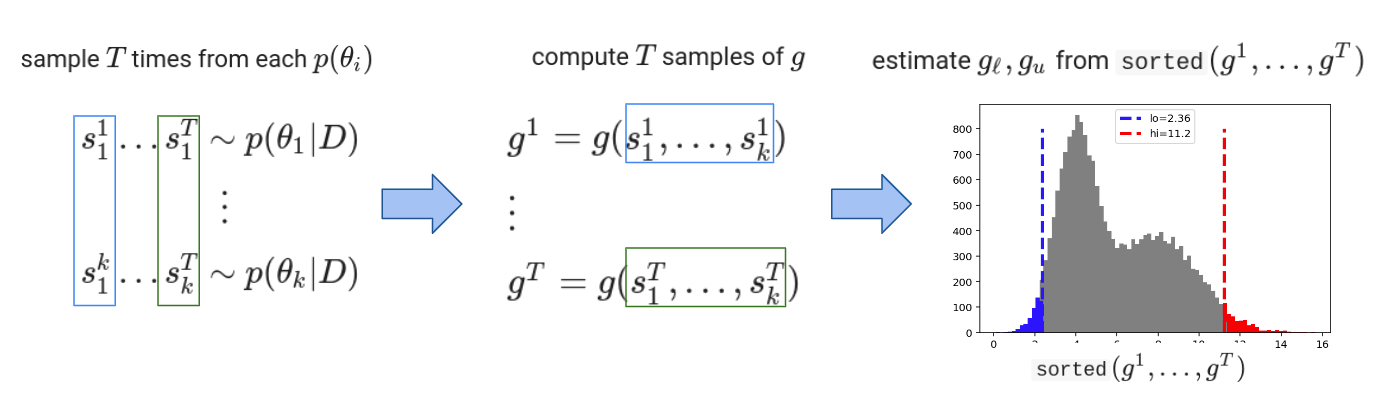}
\end{center}
\vspace{-0.25in}
\caption{Monte Carlo integration to compute confidence intervals for a function $g(\theta_1,\ldots,\theta_k)$, where $\theta_i$'s are
unknown population means and proportions that must be estimated from a sample $D$.} \label{fig:mci}
\end{figure*}

As noted in the introduction, we will be interested in constructing credible intervals over expressions of the form
\begin{equation} \label{eq:mci}
 \int g(\theta_1,\ldots,\theta_k)
      p(\theta_1|D) \ldots p(\theta_k|D)
      d\theta_1 \ldots d\theta_k
\end{equation}
where $g$ is an arbitrary function; each $\theta_i$ is an unknown parameter value that is to be estimated from a dataset $D$; and each $p(\theta_i|D)$ is the posterior over $\theta_i$.  In the cases considered here the construction is correct because the $\theta_i$'s are independent given $D$, generally because they are computed from different subsets of the data.

\emph{Monte Carlo integration} \cite{wasserman2013all} approximates
this integral with bootstrap-like method. 
For each parameter $\theta_i$, we sample $T$ times from its posterior: 
\begin{eqnarray*}
s_1^1,\ldots,s_1^T & = & \mbox{$T$ samples of $p(\theta_1|D)$} \\
 & \vdots & \\
s_k^1,\ldots,s_k^T & = & \mbox{$T$ samples of $p(\theta_k|D)$}
\end{eqnarray*}
We then compute the function $g$ on each of these $T$ samples
\begin{eqnarray*}
g^1 & = & g(s_1^1, \ldots, s_k^1) \\
 & \vdots & \\
g^T & = & g(s_1^T, \ldots, s_k^T)
\end{eqnarray*}
Since $g^1,\ldots,g^T$ is a posterior-weighted sample of $g(\theta_1,\ldots,\theta_k)$,
Eq~\ref{eq:mci} can be approximated by $\frac{1}{N} \sum_{t=1}^T g^t$.  

Now suppose we sort the $g^t$'s to create a long vector $\langle g'^1, \ldots, g'^T \rangle$, and consider the index $t_{\ell} = \lfloor{\alpha T}\rfloor$ for some $\alpha$, say $\alpha=\frac{1}{10}$.  Clearly only a fraction $\alpha$ of the original sample were smaller than $g'_{t_{\ell}}$.  Likewise only a fraction $\alpha$ were larger than $g'_{t_u}$ for $t_u = \lceil(1 - \alpha) T\rceil$, so for a large sample a confidence interval can be easily constructed.  

This process is illustrated in Figure~\ref{fig:mci}.
In the experiments here, we use $T=10,000$ unless otherwise stated, which is large enough that the uncertainty associated with the sampling can be ignored.

\section{Related Work}

\subsection{Past work on PPI}

The difference estimate of Eq~\ref{eq:diff-intro} is well-known in mathematical statistics \cite{difference-est-survey17}, and it is closely related to doubly robust policies in reinforcement learning \cite{doubly-robust-2011}.  Until recently, the difference estimate was not widely known in the AI/ML community, but it was discussed by \cite{doi:10.1126/science.adi6000} as a means of exploiting arbitrary machine learning models as autoraters.

\cite{doi:10.1126/science.adi6000} also discussed a number of generalizations of the difference estimate based on the observation that computing means can be viewed as solving a convex optimization problem; specifically the mean $\mu$ of a sample $Y=y_1,\ldots,y_n$ solves
\[
\mu = \textit{argmin}_{\mu'} \sum_{i} (y_i - \mu')^2
\]
Based on this insight, \cite{doi:10.1126/science.adi6000} propose ``prediction-powered'' algorithms for computing confidence intervals for other statistics of interest, including means, medians, quantiles, and parameters of logistic and linear regressions. 

\cite{doi:10.1126/science.adi6000} also propose methods that apply to solutions to any convex optimization tasks---although these are expensive with the methods of \cite{doi:10.1126/science.adi6000}, requiring grid search over parameter space. 
In later work \cite{angelopoulos2023ppi} propose a more efficient process for finding confidence intervals for general convex optimization methods, and also present a method called \emph{power tuning}, which we discuss below.  Past applications of PPI include ranking chatbots in the Chatbot Arena \cite{boyeau2024autoeval} and evaluating retrieval-augmented generation systems \cite{saadfalcon2024ares}.
In this paper we do not compare to prior PPI experiments constructing confidence intervals for parameters of linear or logistic regression, as we consider prior methods more appropriate for this case.

Technically, the credible intervals produced by Bayesian methods are different from the confidence intervals produced by approaches like that of \cite{doi:10.1126/science.adi6000,angelopoulos2023ppi,boyeau2024autoeval}. In practice, however, both intervals are used in a similar way. We consider the main contribution of this paper over prior work the computation of better intervals, not any perceived benefit of Bayesian interval analysis over classical analysis.

\subsection{Power tuning for PPI} \label{sec:powertune-related}

The difference estimator of Eq~\ref{eq:diff-intro} can be generalized to
\begin{equation} \label{eq:powertune}
\frac{1}{N}\sum_{j=1}^N \lambda f(x_j)
   ~~+~~ \frac{1}{n}\sum_{i=1}^n y_i - \lambda f(x_i)    
\end{equation}  
where $\lambda$ is any constant between 0 and 1 \cite{angelopoulos2023ppi}.  This formulation
is equivalent to the difference estimator when $\lambda=1$, and equivalent to the classical estimator when $\lambda=0$ (and like the standard difference estimator, it has the same expectation as the classical estimator).  Eq~\ref{eq:powertune} thus defines a parameterized family of PPI methods that interpolate between the classical and difference estimate.  \cite{angelopoulos2023ppi} present a closed-form formula for computing $\lambda^*$ that minimizes variance, and call this technique \emph{power tuning}.  The formula for $\lambda*$ makes use of both the labeled and unlabeled data: $\lambda*$ is closely related to the correlation coefficient between $y$ and $f(x)$.  

In this work, we experimentally compare our methods to power tuning, and additionally propose hybrid methods that combine powertuning with novel PPI methods.  
We discuss powertuning and related methods further in Section~\ref{sec:tuning}.

\subsection{Other approaches to re-calibration and ensembles}

One can also view power tuning as re-calibrating $f(x)$, the autorater, by scaling it by the factor $\lambda*$.  More generally, there is a strong similarity between the rectifiers used in PPI and post-hoc calibration of a classifier.   There is a rich literature on post-hoc approaches to calibrating learned classifiers (e.g., \cite{platt1999probabilistic,niculescu2005obtaining}), but there is a clear difference in goals between calibration and PPI: the former is aimed at modifying a learned classifier (the autorater) to make better probabilistic predictions, and the latter is aimed at obtaining better confidence intervals on a model by making use of an existing autorater.

That said, clearly one approach to improving PPI is to use a better-calibrated classifier, perhaps by taking some of the labeled data available and using it for re-calibration.  We leave such approaches as future work here, but we do experimentally explore PPI approaches on both well-calibrated autoraters and poorly calibrated ones.

Power tuning can also be viewed as an ensemble method---a sort of mixture-of-experts approach in which $\lambda$ serves to softly select the classical estimate or the difference estimate.  It is fairly easy to ensemble PPI methods (for instance, by doing multiple-test corrections explicitly) but we again such approaches as future work.

\section{Methods}

\newcommand{\mean}[3]{\textrm{Mean}({#1}:{#2}\in{#3})}
\newcommand{\cprop}[3]{\textrm{Prop}({#1}|{#2}\in{#3})}
\newcommand{\prop}[2]{\textrm{Prop}({#1}\in{#2})}

\subsection{Notation and overview} \label{sec:diff-est}

\subsubsection{Notation} \label{sec:notation}

Following \cite{doi:10.1126/science.adi6000}, we assume 
a sample $S_n = (x_1,y_1), \ldots, (x_n,y_n)$ of $n$ examples drawn from some unknown static distribution, where $y_i$ is a real-valued target output;
a larger sample $\twid{S}_N = \twid{x}_1,\ldots,\twid{x}_N$ for which the target outputs are not available ($N>\!\!>n)$; 
and an \emph{auto-rater function} $f(x)$ which provides a ``good approximation'' of the target $y$ for $x$.  

For example, assume each $x$ is a question/answer pair where the answer is produced by an LLM-based QA system we wish to evaluate, $y$ is a 0/1 gold rating of correctness of the answer, $f(x)$ predicts the human rating on a question/answer pair $x$, and $\expect{y}$ is the accuracy of the QA system on the distribution from which questions are drawn.

Below we write sample means using this notation:
\[ \hat{\mu}^n_y = \mean{y}{y}{{S}_n} \equiv 
    \frac{1}{n} \sum_{(x_i,y_i)\in{}S_n} y_i
\]
with the understanding that this implies a sample standard deviation of 
$\hat{\sigma}_y$; a true (population) mean and standard deviation of $\mu_y$ and $\sigma_y$; a true parameter value $\theta_y^*$; and a posterior $p(\theta_y|{S}_n)$.  The superscript $n$ will be dropped where it is clear from context.  

We denote binomials, estimated from a sample $S=\{a_1,\ldots,a_n\}$, for $a_i\in\{0,1\}$,as
\[ \hat{p}_{A} = \prop{A}{S}
\]
where again $\hat{p}$ implies a corresponding true probability $p_A=\theta^*_{A}$, and a posterior $p(\theta_{A}|S)$.  Estimates for conditional probabilities $p(A=1|B=1)$ for a sample $S=\{(a_1,b_1),\ldots,(a_n,b_n)\}$
are written
\[ \hat{p}_{A|B} = \cprop{A}{B}{S}
\]
and estimates of $p(A=1|B=0)$ will be written $\hat{p}_{A|\neg B}$ for conciseness. 
When $A$ or $B$ is a multinomial then we use a similar notation, without abbreviating $P(A=1)$ to $P(A)$ or $P(A=0)$ to $P(\neg A)$.

All of these quantities are modeled as random variables that depend on data, and each of them has a prior and posterior, so we can meaningfully write things like $p(\hat{\mu}^n_y | D)$.  

\subsubsection{Overview: Designing a Bayesian PPI method}

We assume we are given is a statistic, the \emph{target estimand}, we want to measure in expectation.  The target estimand is denoted $e(S_n,\twid{S}_N)$. 

The first step is
to introduce a second statistic, the \emph{proxy estimand}, that has the same expectation as $e$ but lower variance. The equality of expectations is easy to verify for the cases considered here.  The proxy estimand is denoted $g(S_n,\twid{S}_N)$, and is a function of random variables.  In this paper, these random variables are all means or proportions, derived from the data $D$, with known posteriors, so in the general case $g$ can be written
\[
g(S_n,\twid{S}_N) = h(\theta_1, \ldots, \theta_k)
\]

Next, we choose a posterior $p(\theta_i|D)$ for each random variable in $g$.  The priors we use in this paper are weak, uniformed conjugate priors, as discussed above and detailed in Sec~\ref{sec:python}: a Gaussian or Student's $T$ distribution for means, a Beta for simple proportions (binomials), and a Dirichlet for multinomials.

Finally, we measure variance of $g$ and compute a confidence interval $\ell_g, u_g$ using Monte Carlo integration, as described in Section~\ref{sec:mci}.

\subsection{The Bayesian difference estimate}

For the difference estimate we begin with the target estimand
\[  e(S_n,\twid{S}_N, f) \equiv \mean{y}{(x,y)}{S_n}
\]
The proxy estimand is
\[  g(S_n,\twid{S}_N, f) \equiv  \hat{\mu}^N_{f(x)} + \hat{\mu}^m_{y-f(x)}
\]
where
\begin{eqnarray*}
    \mu^N_{f(x)} & \equiv & 
    \mean{f(x_j)}{x_j}{\twid{S}_N}\\
    \mu^n_{y-f(x)} & \equiv & 
    \mean{y - f(x_i)}{(x_i, y_i)}{{S}_n}
\end{eqnarray*}

It is straightforward to verify that 
$\expect{g}=\expect{e}$, and we use the usual priors.
The Bayesian difference estimate, as well as the other PPI methods described here, is summarized in Table~\ref{tab:methods}.

This variant of the difference estimate is presented for pedagogical purposes, not practical ones.  Experimentally, it is virtually identical in performance to the approach of \cite{doi:10.1126/science.adi6000}---although the supporting theory is different, experimentally it gives essentially the same confidence intervals.  Hence experiments below that use the difference estimate as a baseline generally use the method of \cite{doi:10.1126/science.adi6000}.

\subsection{Stratified estimates}

With the priors selected above, the Bayesian difference gives the same confidence intervals as the traditional method. 
Our first novel PPI method, stratified estimates, is based on the observation that, 
while the difference estimate works best when the variance of 
$\mean{y - f(x)}{(x,y)}{S_n}$ is small, this can be true in two cases: (1) if $f(x)\approx y$, but also when (2) $f(x) \approx y + b$, where $b$ is constant.  In other words, it is not necessary for the autorater to be unbiased, as long as its bias is \emph{consistent} across examples.

There are many cases, however, where autorater bias is not consistent.  Consider, for example, a task for which humans give a ``star rating'' which is an integer between 1 and 5, and an autorater is trained to predict that rating.  If the human scores are frequently extreme (i.e., 1 or 5, and rarely 2, 3 or 4), and the autorater is trained to minimize loss, it may well trend low on examples a human would rate 5 stars, and high on examples a human would rate as 1 star.  In this case you might see two regimes of autorater bias: perhaps when $f(x) > 2.5$, then $y \approx f(x) + 1$, and when $f(x) \leq 2.5$ then $y \approx f(x) - 1$.

One way of adjusting for this effect would be create a \emph{partition} function $\pi$, which maps $f(x)$ to a discrete set of intervals, and then construct a difference estimate over each interval.  In this example, we would define
\[
\pi(f(x)) = \left\{ \begin{array}{ll}
                     \textit{lo} & \mbox{if $f(x) > 2.5$} \\
                     \textit{hi} & \mbox{else}
                    \end{array}
            \right.
\]
Let us define 
\begin{eqnarray*}
S_n^\textit{lo} & = \{(x_i,y_i)\in S_n : \pi(f(x_i)=\textit{lo} \\   
S_n^\textit{hi} & = \{(x_i,y_i)\in S_n : \pi(f(x_i)=\textit{hi} \\   
\end{eqnarray*}
and define $\twid{S}_N^\textit{lo}$ and $\twid{S}_N^\textit{hi}$ similarly.  Consider the random variables
\begin{eqnarray*}
\hat{\mu}^{\textit{lo}}_{f(x)} & = & 
     \mean{f(x)}{x}{\twid{S}_N^\textit{lo}} \\
\hat{\mu}^{\textit{hi}}_{f(x)} & = & 
     \mean{f(x)}{x}{\twid{S}_N^\textit{hi}} \\
\hat{\mu}^{\textit{lo}}_{y-f(x)} & = & 
    \mean{y - f(x)}{y}{S_n^\textit{lo}} \\
\hat{\mu}^{\textit{hi}}_{y-f(x)} & = & 
    \mean{y - f(x)}{y}{S_n^\textit{hi}} \\
\hat{p_\textit{lo}} & = & 
     \prop{\pi(f(x))=\textit{lo}}{\twid{S}_N}
\end{eqnarray*}
The proxy estimate $g(S_n, \twid{S}_N)$ is
\[
( \hat{\mu}^\textit{lo}_{f(x)} + \hat{\mu}^\textit{lo}_{y-f(x)}) \cdot \hat{p}^\textit{lo} 
+ ( \hat{\mu}^\textit{hi}_{f(x)} + \hat{\mu}^\textit{hi}_{y-f(x)}) \cdot (1 - \hat{p}^\textit{lo})
\]
In general, if there are $K$ partitions, the model is a weighted sum of $K$ difference estimates, where the difference estimate for partition $k$ is constructed using the labeled and unlabeled data mapped to partition $k$,
and the weight of that estimate is the fraction of (unlabeled) data mapped to partition $k$.  It can easily be shown that the 
expectation of $g$ is still the population mean.   The general form of the stratified difference estimate is given in Table~\ref{tab:methods}.

This \emph{stratified difference estimate} essentially it combines PPI methods with stratified sampling \cite{Singh1996},
and like standard stratified sampling, it can be used with any partitioning scheme.  In the experiments below we explore two partitioners.  In the simple case, we divide $x\in\twid{S}_N$ into $K$ equal-population bins.  We also explore using the labeled data in $S_n$ to build a regression tree \cite{lewis2000introduction}, and taking the leaves of the tree as partitions: see~\ref{sec:reg-tree} for details.

\subsection{Chain rule estimates}

\subsubsection{A simple chain rule estimate}

We now consider extending the difference method to discrete autoraters---for instance, autoraters based on prompted LLMs.  For this new method we denote the human rating with the random variable $H$ and the autorater's rating as $A$, so
\begin{eqnarray*}
    S_n & = & \{ (a_1, h_1), .... (a_n, h_n) \} \\
    \twid{S}_N & = &  \{ a_1, .... h_N \}
\end{eqnarray*}
Notice that $H$ and $A$ are dependent on a randomly chosen question/answer pair, $x$.  We don't use $x$ below, but to ensure parameter independence, the $x$'s associated with $S_n$ and $\twid{S}_N$ should be non-overlapping.

Our target estimand is the expected human rating:
\[
e(S_n, \twid{S}_N) = \prop{H}{S_n}
\]
and the proxy estimand uses the chain rule to evaluate it:
\[
 g(S_n,\twid{S}_N) \equiv 
    \hat{p}_{H|A} \cdot \hat{p}_{A} +
    \hat{p}_{H|\neg A} \cdot ( 1 - \hat{p}_{A})
\]
where
\begin{eqnarray*}
    \hat{p}_A & = & \prop{A}{\twid{S}_N} \\
    \hat{p}_{H|A} & = & \cprop{H}{A}{{S}_n} \\
    \hat{p}_{H|\neg A} & = & \cprop{H}{\neg A}{{S}_n}
\end{eqnarray*}
By analogy with the difference estimator, we call this method a \emph{chain rule estimate}.
It is clear that $\expect{g}=\expect{e}$, but it is less obvious why this trick should reduce variance.  However, 
recall the standard error of a binomial with probability $p^*$ estimated from $n$ samples is approximately $\sqrt{p^*(1-p^*)/n}$.
Examining the proportions in $g$, we see that
$\hat{\sigma}_{A}$ is small since it is computed from $N$ samples, and $N$ is large. If the autorater is accurate, then $p^*_{H|\neg A}$ and $(1 - p^*_{H|A})$ will be small, so $\hat{\sigma}_{H|\neg A}$ and $\hat{\sigma}_{H|A}$ will be small.

This model is quite similar to the stratified estimate discussed above, as they both marginalize over different cases. One technical difference is that it uses binomials instead of difference estimates in the ``inner loop'' of the marginalization sum.  It is also arguably a clearer description of how to make use of discrete autorater values.

\newcommand{\ediffest}{$\mean{y}{y}{{S}_n}$}
\newcommand{\gdiffest}{$\hat{\mu}^{N}_{f(x)} + \hat{\mu}^{n}_{y-f(x)}$}
\newcommand{\muf}{$\hat{\mu}^{N}_{f(x)} = \mean{f(x)}{x}{\twid{S}_N}$}
\newcommand{\mudiffs}{\multicolumn{1}{l}{$\hat{\mu}^{n}_{y-f(x)} = $}}
\newcommand{\mudiffe}{\multicolumn{1}{l}{~~~$\mean{y - f(x)}{(x,y)}{S_n}$}}

\newcommand{\epart}{$\mean{y}{y}{{S}_n}$}
\newcommand{\gpart}{ \( %
 \sum_{i=1}^{K} (\hat{\mu}^{i}_{f(x)} + \hat{\mu}^{i}_{y-f(x)}) \cdot %
  \hat{p}_{i} \)}
\newcommand{\forpart}{for $i=1,\ldots,K$:}
\newcommand{\ppart}{~~$\hat{p}_{i} = \prop{\pi(f(x))=i}{\twid{S}_N}$}
\newcommand{\mupartf}{~~$\hat{\mu}^{i}_{f(x)} = \mean{f(x)}{x}{\twid{S}_N^i}$}
\newcommand{\muparts}{\multicolumn{1}{l}{~~$\hat{\mu}^{i}_{y-f(x)} = $}}
\newcommand{\muparte}{\multicolumn{1}{l}{~~~~$\mean{y - f(x)}{(x,y)}{S_n^i}$}}

\newcommand{\epd}{$\prop{H}{S_n}$}
\newcommand{\gpd}{$\hat{p}_{H|A} \cdot \hat{p}_{A} + 
    \hat{p}_{H|\neg A} \cdot ( 1 - \hat{p}_{A})$}
\newcommand{\pA}{$\hat{p}_A  = \prop{A}{\twid{S}_N}$}
\newcommand{\pHgA}{$\hat{p}_{H|A} =  \cprop{H}{A}{{S}_n}$}
\newcommand{\pHgnA}{$\hat{p}_{H|\neg A} =  \cprop{H}{\neg A}{{S}_n}$}

\newcommand{\eapd}{\epd}
\newcommand{\gapdf}{ \( %
  \sum_{a \in \{y,n,u\}} \hat{p}_{H|A=a} \cdot \hat{p}_{A=a} \) }
\newcommand{\pAfor}{for $a \in \{y,n,u\}$: }
\newcommand{\pAp}{ \( %
   ~~\hat{p}_{A=a} = \prop{A=a}{\twid{S}_N} \)}
\newcommand{\pAm}{ \( %
   ~~\hat{p}_{H|A=a} = \cprop{H}{A=a}{{S}_n} \)}

\newcommand{\epairs}{$\prop{H=w}{S_n} ~ - ~$}
\newcommand{\epaire}{$\prop{H=l}{S_n}$}

\newcommand{\gpairs}{$\sum_{a \in \{w,l,t\}} \hat{p}_{H=w|a} \cdot \hat{p}_{A=a} ~ -$}
\newcommand{\gpaire}{$\sum_{a \in \{w,l,t\}} \hat{p}_{H=l|a} \cdot \hat{p}_{A=a}$}
\newcommand{\pApaira}{for $a\in\{w,l,t\}$, $h\in\{w,t\}$:}
\newcommand{\pApairb}{~~~$\hat{p}_{A=a} = \prop{A=a}{\twid{S}_N}$}
\newcommand{\pApairc}{~~~$\hat{p}_{H=h|a} = \cprop{H=h}{A=a}{\twid{S}_N}$}

\begin{table*}
\begin{small}
\begin{tabular}{cclc}
Estimand                &  Designed Statistic   & Means or Proportions Used in $g$  &   Comments        \\ 
$e(S_n,\twid{S}_N)$     & $g(S_n,\twid{S}_N)$   &   &  \\
\hline \\
\ediffest               & \gdiffest             & \muf      &  difference \\
                        &                       & \mudiffs  &  estimate \\
                        &                       & \mudiffe  &  \\
\\ \hline \\

\epart                  & \gpart                & \forpart   & stratified \\
                        &                       & \ppart    & difference \\
                        &                       & \mupartf  & estimate \\
                        &                       & \muparts  &  \\
                        &                       & \muparte  &  \\
\\ \hline \\

\epd                    & \gpd                  & \pA       & chain rule   \\
                        &                       & \pHgA     & estimate  \\
                        &                       & \pHgnA    & \\
\\ \hline \\

\eapd                   & \gapdf             &  \pAfor     & chain rule \\
                        &                    &  \pAp     & estimate with \\
                        &                       & \pAm    & an abstaining \\
                        &                       &    & autorater \\
                        &                       &     & \\

\\ \hline \\

\epairs                 & \gpairs               &  \pApaira         & chain rule  \\
\epaire                 & \gpaire               &  \pApairb         & estimate for \\
                        &                       &  \pApairc         & paired tests \\

\\ \hline
\end{tabular}
\end{small}

\caption{Overview of PPI methods used in this paper.  The difference estimate
is a Bayesian version of prior work and the remaining models are novel.} \label{tab:methods}
\end{table*}

\subsubsection{PPI for abstaining autoraters} \label{sec:abstain-method}

There are a number of natural extensions to the chain rule estimate, and we consider two of these in depth in this paper.  One is for autoraters that do not always give a yes-or-no answer.  It might be that the autorater output cannot be parsed as expected, or it might be that the autorater is designed to answer ``unknown'' when presented with a question/answer pair whose correctness cannot be determined.  To model this, we can assume that $A$ is a multinomial random variable with three outputs, $y$ for ``acceptable'', $n$ for ``not acceptable'', and $u$ for ``unknown''.  If we assume the gold human labels are still binary, we can modify our model as follows:
\begin{eqnarray*}
    \hat{p}_{A=a} & = & \prop{A=a}{\twid{S}_N}, ~~a \in \{y,n,u\} \\
    \hat{p}_{H|A=a} & = & \cprop{H}{A=a}{{S}_n}, ~~a \in \{y,n,u\}  \\
 g(S_n,\twid{S}_N) & = &  
    \sum_{a \in \{y,n,u\}} \hat{p}_{H|A=a} \cdot \hat{p}_{A=a}
\end{eqnarray*}
In this case, we don't expect the variance of $\hat{p}_{H|0}$ to be especially small, so it should also be the case that the  the autorater not abstain too often---i.e., that $p(A=u)$ is small.

\subsubsection{PPI for side-by-side tests} \label{sec:sxs}

Another common kind of labeling is variously called a side-by-side test, an A/B test, or a paired test.  In this case raters are asked to compare two alternative outputs for the same input string, and either express a preference between them, or say that the outputs are equally good.
In Section~\ref{sec:sxs-results}, we describe a novel estimate for this based on the chain rule estimate.  For completeness, this estimate is also shown in Table~\ref{tab:methods}.

\section{Case studies}


\subsection{Continuous autorater predictions and stratified estimators}


\newcommand{\bst}[1]{\textbf{#1}}
\newcommand{\ours}{\textit{ours:}}

\begin{table*}[tb]
\begin{center}
\begin{tabular}{lrrrr|rrrr}
\toprule
 &  \multicolumn{4}{c}{CI Width}
 & \multicolumn{4}{c}{CI Ratio to Classical} \\
 Method & Seahorse1 & Seahorse2 & AQA1 & AQA2 & Seahorse1 & Seahorse2 & AQA1 & AQA2 \\
\midrule
 & \multicolumn{8}{c}{\textit{methods that do not tune hyperparameters on $S_n$}} \\
\midrule
classical & 0.112 & 0.107 & 0.108 & 0.112  & 1.000 & 1.000 & 1.000 & 1.000\\
difference estimate & 0.098 & 0.104 & 0.099 & 0.096 & 0.873 & 0.968 & 0.914 & 0.855\\
\ours{} chain rule & 0.100 & 0.102 & 0.094 & 0.094  & 0.892 & 0.951 & 0.870 & 0.833 \\
\ours{} stratified($K=5$) 
& \bst{0.094} & \bst{0.100} & \bst{0.084} & \bst{0.080}  
& \bst{0.839} & \bst{0.929} & \bst{0.778} & \bst{0.708}  \\
\midrule
 & \multicolumn{8}{c}{\textit{methods that tune on $S_n$}} \\
\midrule
tree(5) & 0.093 & 0.100 & 0.082 & 0.079 & 0.829 & 0.932 & 0.765 & 0.703 \\
stratified($K=*$) & 0.092 & 0.099 & 0.082 & 0.079  & 0.826 & 0.919 & 0.759 & 0.701 \\
difference estimate + ptune  & 0.095 & 0.100 & 0.089 & 0.089 & 0.848 & 0.928 & 0.825 & 0.788\\
stratified($K=5$) + ptune  & 0.093 & 0.099 & 0.082 & 0.078 & 0.826 & 0.916 & 0.764 & 0.688 \\
stratified($K=*$) + ptune & 0.092 & 0.098 & 0.081 & 0.077 & 0.822 & 0.914 & 0.748 & 0.685 \\
tree($K=5)$ + ptune
& \bst{0.091} & \bst{0.097} & \bst{0.078} & \bst{0.075} 
& \bst{0.808} & \bst{0.902} & \bst{0.726} & \bst{0.668} \\
\bottomrule
\end{tabular}
\end{center}
\caption{CI widths and CI width ratio for all datasets with $n=300$ human-labeled examples.  Methods marked with ``+ ptune'' use powertuning \cite{angelopoulos2023ppi} within partitions. 
The difference estimate baseline is implemented as in \cite{doi:10.1126/science.adi6000}.  All methods that include tuning are novel (``ours'') except for the difference estimate with power tuning (called PPI++ in \cite{angelopoulos2023ppi}).}
\label{tab:n300}
\end{table*}

\begin{figure*}[t]
\begin{center}

\begin{tabular}{cc}
 \includegraphics[width=0.35\textwidth]{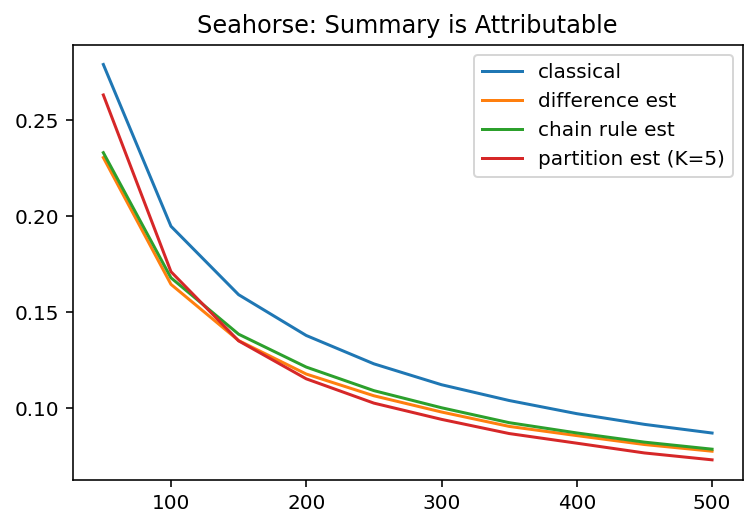} & 
 \includegraphics[width=0.35\textwidth]{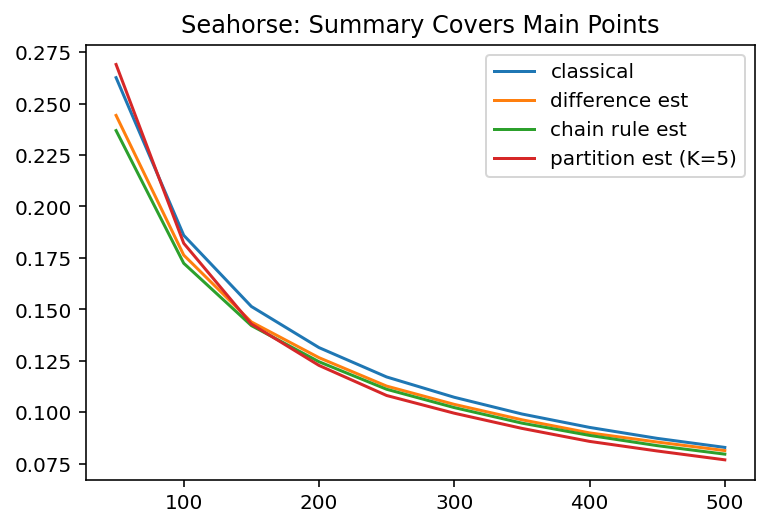} \\
 \includegraphics[width=0.35\textwidth]{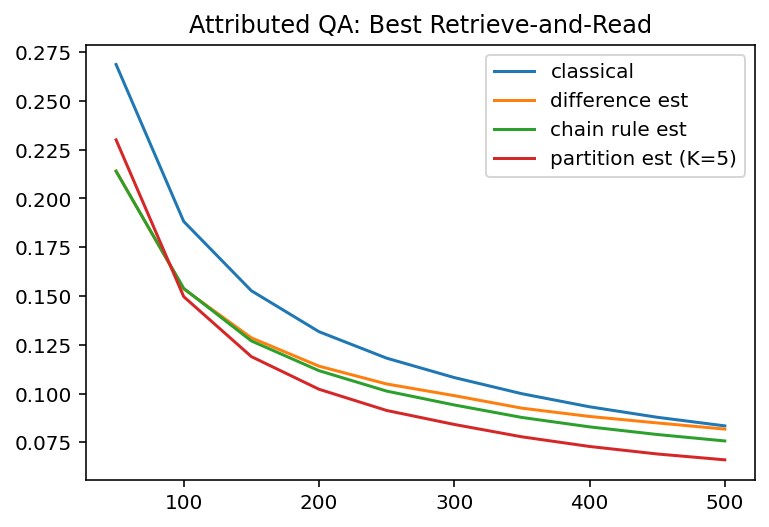} & 
 \includegraphics[width=0.35\textwidth]{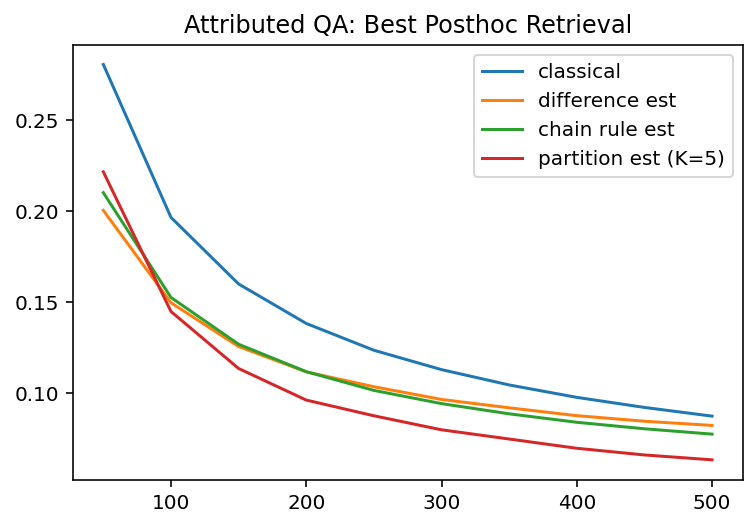} \\
\end{tabular}
\end{center}
\caption{Comparing stratified estimates on multiple datasets.  On all graphs, the $x$ axis is number of human-labeled examples $n$, and the $y$ axis is confidence interval width.  All PPI methods improve over classical approaches.  The chain rule estimate and difference estimate are generally comparable, and the stratified estimate improves performance substantially over each of them (see Section~\ref{sec:impact-part}).}
\label{fig:part}
\end{figure*}

\begin{table}[htb]
    \centering
    \begin{tabular}{lrrrr}
\toprule
 Method & SH1 & SH2 & AQA1 & AQA2 \\
 \midrule
 classic & 96.6 & 96.5	& 97.8	& 97.3 \\
 difference est & 96.9 &95.5 & 98.8	& 99.0 \\
 chain rule & 97.4 & 96.4 & 98.3	& 98.9 \\
 strat($K=5)$ & 96.3 & 95.8 & 97.3	& 97.0 \\
 \midrule
 tree($K=5$) & 96.3 & 95.7 & 96.3 & 95.5 \\
 tree($K=*$) & 92.2 &  88.5 & 91.3 & 89.4 \\
 strat($K=*$) & 96.9 & 95.7 & 97.6 & 96.9 \\
 \multicolumn{5}{l}{\textit{$+$ powertune}} \\
 difference est & 97.6 & 95.9 & 99.3 & 98.5 \\
 strat($K=*$) & 96.8 & 95.7 & 97.5 & 96.4 \\
 tree($K=5$ & 96.3 & 95.8 & 95.5 & 95.1 \\
 tree($K=*)$ & 89.7 & 87.7 & 87.7 & 86.1 \\
 \bottomrule
    \end{tabular}
    \caption{Coverage results at $n=300$ for 1000 trials of each method. All methods, even the tuned methods, obtained coverage at least as large as the expected 95\% except for those that combine regression-tree learning with search over the optimal number of partitions.}
    \label{tab:coverage-300}
\end{table}

\subsubsection{Datasets}

A common task in NLP is evaluating long-form outputs from LLMs: for instance, the outputs of chatbots, summarizers, and systems that provide long answers to information-seeking questions.  In general this is done with human ratings, and for many tasks, autoraters also exist.  In this setting, there will be a limited set of gold ratings $x,z,y$,
and two LLM models: a \emph{generator model} which produces answer an output $z$ from an input prompt $x$, and a \emph{autorater model} which takes a pair $x,z$ and outputs a predicted human rating $\hat{y}$.

As the first testbed for these tasks, we use data from the Seahorse project \cite{clark-etal-2023-seahorse}.  In this work, the authors considered generator models that summarize a source document $x$ in a multilingual setting.  This is a compelling task since obtaining labels for multiple languages is expensive.  Seahorse also includes labels for many languages, many systems, and many dimensions of summary quality: overall there are labels for 6 dimensions of summary quality, for 9 summarization systems, and 6 languages.  

We consider here two quality dimensions, one which captures if the summary is fully attributable to the source document, and one which captures if the summary captures the main idea of the source.  (These are the dimensions for which there was the greatest variance between different summarization models. )  We consider one summarization system---a finetuned 13B parameter multilingual mT5 \cite{xue2020mt5} model, the strongest model based on an open-source LLM.  The judge models for each dimension are also mT5-XXL finetuned models, which output probability scores.  The data contains 2728 examples for these two task, all of which have both human ratings and autoratings.  

As a second testbed we used data distributed by the authors of \cite{bohnet2023attributed}, which compared many models on the task of \emph{attributed question answering}.  In attributed question answering, the goal of the QA system is to output an answer and a document that supports that document---i.e., a generation $z$ is a pair $(a_x,d_x)$ where $a_x$ is an answer to question $x$, and $d_x$ is a document $d_x$ that supports the answer, and the generation is considered correct $(y=1)$ if the answer is indeed supported by the document.  We evaluated two ``generator'' models from \cite{bohnet2023attributed}: the highest-scoring ``retrieve-and-read'' QA system\footnote{RTR-10 in the dataset. Retrieve-and-read (RAG) models use a dense retriever with $x$ as query to find candidate documents $d_x$ from a corpus and then generate an answer $a_x$ with a fine-tuned LLM that takes $d_x$ as context.}, and the highest-scoring ``post-hoc retrieval'' system.\footnote{Post-6 in the dataset.  Post hoc retrieval systems generate an answer $a_x$ and then search for a supporting document $d_x$ using $a_x$ as a query.}  
The autorater for this task is a natural-language entailment model: a 11B parameter T5 \cite{JMLR:v21:20-074} model fine-tuned on a collection of NLI-related tasks \cite{honovich2022true}, which again gives probability scores.  These datasets have 1000 human labels and 3000 autorater labels.

\subsubsection{Initial experimental results}

Following \cite{doi:10.1126/science.adi6000} we evaluate PPI method performance using the \emph{width} of a confidence/credible interval. Using $\ell_e, u_e$ for the classical confidence interval for the estimand $e$ and $\ell_g,u_g$ for the PPI confidence interval of the designed statistic $g$, the \emph{width of the classical confidence interval} is $u_e - \ell_e$, and similarly for the PPI interval.  We also consider the \emph{CI width ratio}, which we define as $\frac{u_g - \ell_g}{u_e - \ell_e}$, i.e., the CI width of a PPI test normalized by the width of the classical test.

We compare the classical estimate, which uses only the human labels $S_n$ to a stratified model with $K=5$ equal-frequency partitions.  We did not tune the number of partitions at all in these initial experiments.  We also compare to the difference estimate (called simply PPI in the experiments of \cite{angelopoulos2023ppi,doi:10.1126/science.adi6000}), and, for completeness, the chain rule estimate.  The chain rule estimate is designed for discrete autoraters, and was used here by discretizing the autorater scores using a threshold of 0.5.    The difference estimate baseline is the classical method \cite{doi:10.1126/science.adi6000}, not the Bayesian version described Table~\ref{tab:methods}.

The top rows of Table~\ref{tab:n300} shows these results, averaged over 1000 trials each.

We also use the  experimental procedure of \cite{doi:10.1126/science.adi6000} to simulate the performance of statistical tests with varying numbers of human labels $n$.  We pick 10 different numbers $n$ ranging from 100 to 500, and for each value of $n$, we sample $n$ of the cases with human labels for the human-labeled set $S_n$, and use all the remaining autorater labels as $\twid{S}_N$.  We repeat this 100 times for each value of $n$, and graph the average CI width over all trials for 95\% confidence intervals.  These results are shown in Figure~\ref{fig:part}.

As shown in Table~\ref{tab:n300} at $n=300$, the difference estimate performs surprisingly well, and is comparable and to the chain rule estimate perform similarly on all dataset, despite not having access to the autorater's actual numeric scores.  As expected, the stratified estimate gives a noticeable improvement over the others. 

There is a larger improvement for all PPI methods in the attributed QA tasks: note that one important difference between the Seahorse data and the attributed QA data is that English-language entailment is a well-studied problem with large amounts of training data available, so one would expect the autorater here to be strong.  

Figure~\ref{fig:part} adds nuance to this view, showing that the stratified methods performs less well comparatively with small $n$, but improves over other methods substantially for larger $n$.  This makes sense, because there will be variance associated with the rectifier in each partition, and this variance will be larger when the partition has few human-labeled examples.  Additional experiments exploring this are in Appendix~\ref{app:seahorse}.  
The crossover between the stratified estimate with $K=5$ and the difference estimate is between $n=100$ and $n=150$ for these datasets.

\subsubsection{Practical implications} \label{sec:impact-part}

The graphs of Figure~\ref{fig:part} make it fairly easy to see the practical impact of the improved test.  As an example, drawing a horizontal line on the lower right-hand graph at around $y=0.10$ shows that similar confidence sized confidence intervals using 200 examples of the stratified estimate as using 400 examples of the classical estimate, so in this case labeling costs could be potentially reduced by half.  On the upper right-hand graph, where the differences between methods are smaller, a line in the same $y$ position shows that 300 examples with the stratified estimate are as useful as around 360 labels for the classical method, a 20\% reduction in labeling.

\subsubsection{Coverage tests}


The other important property for a statistical test to have, of course, is that an interval be calibrated: e.g., that a 95\% interval contain the true value of the parameter at least 95\% of the time.  Since for each of these tasks, a fairly large human-labeled dataset is available, we used the mean value obtained using all the human-labeled data as a proxy for the ``true'' value of the unknown parameter $\theta^*$, an measured test \emph{coverage}, i.e., how often the interval given by each test contains this $\theta^*$.  For a 95\% confidence interval, this should occur 95\% of the time.

In Figure~\ref{fig:part}, we aggregate all the tests for each method and each task, so there are a total of sixteen cases, with 1000 intervals considered for each case.  In each case the coverage was 95\% or greater, as expected, with the minimum coverage estimate being 95.5\% (for the difference estimate on the ``Main Points'' Seahorse dimension).
In conducting the experiments for Figure~\ref{fig:vary-part-k}, the stratified estimate was run 1000 times each for 4 values of $K$.  In each of the four cases the coverage was 97.2\% or greater, confirming the theoretical predictions.


\subsubsection{Tuning and powertuning PPI estimates} \label{sec:tuning}

In statistical testing, one must be wary of conducting multiple tests and choosing the best, as this can lead to results that are not well-calibrated---i.e., the true parameter value will lie outside the suggested confidence interval more frequently than expected.  This is why in the experiments of Figure~\ref{fig:part} we used a fixed $K=5$ for the number-of-partitions hyperparameter.   

In many practical settings, however, it is necessary to evaluate systems on many related tasks: e.g., one consider scaling the Seahorse data from 9 languages to 90, or evaluating dozens of variants of a chatbot system during development.  In such settings it is certainly plausible to tune a PPI estimate on a subset of the available tasks and run the well-tuned task on the others, eliminating any multiple-test worries.  Past experiments of \cite{angelopoulos2023ppi} also show that power tuned models seem to be well-calibrated even when the $\lambda^*$ parameter is set by (re-)using $S_n$.
(Recall from Section~\ref{sec:powertune-related} that \emph{powertuning} generalizes the difference method to
\[
 g(S_n, \twid{S}_N) 
   = \frac{1}{N}\sum_{j=1}^N \lambda f(x_j)
   ~~+~~ \frac{1}{n}\sum_{i=1}^n y_i - \lambda f(x_i)    
\]
for any $\lambda$ and also provides a method for picking a $\lambda*$ to minimize variance, thus finding the best test in a family.)

In summary, even if some gaps remain in our formal understanding of the practice of tuning PPI estimates with the labeled data, it seems important to consider these techniques.  In this section we consider methods for tuning hyperparameters of these methods.

\label{sec:reg-tree}

Tuning $K$ is conceptually easy---one can simply run the method with different values of $K$ and adopt the $K^*$ that gives the smallest variance. More generally, the way in which partitions are constructed could also be varied, by intelligently picking partitions to be as large as possible subject to the constraint that $y-\hat{y}$ stays roughly constant within partitions.  One easy-to-implement version of this idea is to fit a regression tree \cite{lewis2000introduction} to the labeled data, using the autorater score $\hat{y}$ as an input feature and the human label $y$ as the target value, and then use the leaves of the tree as partitions.

The bottom rows of Table~\ref{tab:n300} presents results for a variety of tuned PPI methods.  We first present the regression-tree partitioning method, constrained to output a tree with at most $K=5$ leaves, and varying the number of partitions $K$ for the equal-frequency partitions with and without powertuning (see Appendix~\ref{app:seahorse} for details.).  We then consider adding powertuning to all methods (except the chain rule.)  Many of these methods lead to substantially improved results, with the powertuned regression-tree method performing best.

\subsubsection{Coverage Tests}

Without a strong theory that predicts coverage for tuned PPI methods, experiments on coverage are essential. 
The results for each result in Table~\ref{tab:n300} are averaged over 1000 trials each, and we recorded for each of these trials the coverage of the method (i.e., how often the CI contains the true value, as estimated from all available human-labeled examples.)  The results are shown in Table~\ref{tab:coverage-300}.  Interestingly, all of the methods, including those that use tuning, are still well-calibrated, \emph{except} for the ones that combine tree learning and search over $K$ (shown as ``tree($K=*$)'' in the table.)  The poor coverage results for these methods are why their CI widths are not presented in Table~\ref{tab:n300}.

\subsection{Uncalibrated autoraters and chain-rule estimates} \label{sec:openqa}

The previous section considers autoraters that are based on well-calibrated, fine-tuned models.  A common recent practice is to use prompted models for autoraters, which we consider next.  Note that some of the strategies available for fine-tuned models cannot be used here: in particular, since only a few different predictions are made by autoraters, partitioning cannot be used effectively.

\subsubsection{Datasets}

In a recent paper \cite{kamalloo-etal-2023-evaluating} showed that the most commonly-used metrics for evaluating open-book QA systems are very biased, relative to human raters.  Among other results, they evaluated 10 widely-used open-book QA systems on 3600 examples with automatically-computed metrics, such as exact-match or token-F1 similarity to known answers, and also evaluated the same systems with human raters on a smaller set of 300 examples.  The human ratings are very different (see Table~\ref{tab:kamalloo-motivate}), showing gains in accuracy of more than 20\% in several cases.
The authors discuss potential solutions, including using prompted LLMs as autoraters, but ultimately conclude ``At this time, there appears to be no substitute for human evaluation [of open-book QA systems].''

\begin{table}
\centering
\begin{tabular}{l|rr|rr}
\toprule
     & EM & F1 & Human & (delta) \\
     & \multicolumn{2}{c}{3.6k examples} & \multicolumn{2}{|c}{300 examples} \\
\midrule
DPR &  40.9 & 47.8 & 58.8 & +12.9 \\
FiD & 46.5 & 53.7 & 64.8 & +17.0 \\
ANCE/FiD & 47.3 & 54.8 & 65.8 & +17.6 \\
RocketQAv2/FiD  & 47.7 & 55.6 &  70.1 & +20.3 \\
Contriever/FiD  & 47.9 & 55.4 & 66.5 & +20.0 \\
FiD-KD  & 49.6 & 57.4 &  73.1 & +22.3 \\
GAR+/FiD & 49.8 & 57.4 & 69.4 & +18.2 \\
EviGen  & 49.8 & 57.0 & 67.1 & +15.3 \\
EMDR2  & 51.5 & 59.5 &  73.1 & +19.9 \\
R2-D2  & 52.4 & 59.0 &  71.4 & +18.6 \\
\bottomrule
\end{tabular}
\caption{Human vs automatic labels for open-book QA systems, from \cite{kamalloo-etal-2023-evaluating}.}
\label{tab:kamalloo-motivate}
\end{table}

We experimented with the data released by \cite{kamalloo-etal-2023-evaluating}. The data includes model outputs for ten models on 3600 examples, but only includes human labels for the smaller set of 300 examples.  Since PPI methods require a larger set of autorater examples, we implemented an prompted autorater using Gemini Pro \cite{team2023gemini}, and ran this autorater on the the question/answer pairs for every model on each of the 3.6K standard NQ test examples, resulting in 300 human labels and 3.3k autorater labels for each model.  Similar to the autoraters of \cite{kamalloo-etal-2023-evaluating}, our autorater takes a question/answer pair, plus the set of known short answers, and predicts if the answer is acceptable or not.  No probability scores are associated with an autorating.

We elected to evaluate performance at $n=300$ labeled examples exclusively, rather than constructing curves as in Figure~\ref{fig:part}.

\subsubsection{An abstaining, uncalibrated autorater} \label{sec:abstain}

Our model is prompted to produce binary values, but occasionally outputs results that are cannot be easily parsed into a correct/incorrect decision.  A principled way of dealing with this was discussed in Section~\ref{sec:abstain-method}: we use the chain rule estimater, and model the LLM as producing an ``unknown'' value as a third possible outcome for the random variable $A$, which now could take on the values ``y'' (for an acceptable question/answer pair), ``n'' (for an unacceptable pair), or ``u'' for an uncertain result. 


\begin{figure}[t]
\begin{center}
    \includegraphics[width=0.45\textwidth]{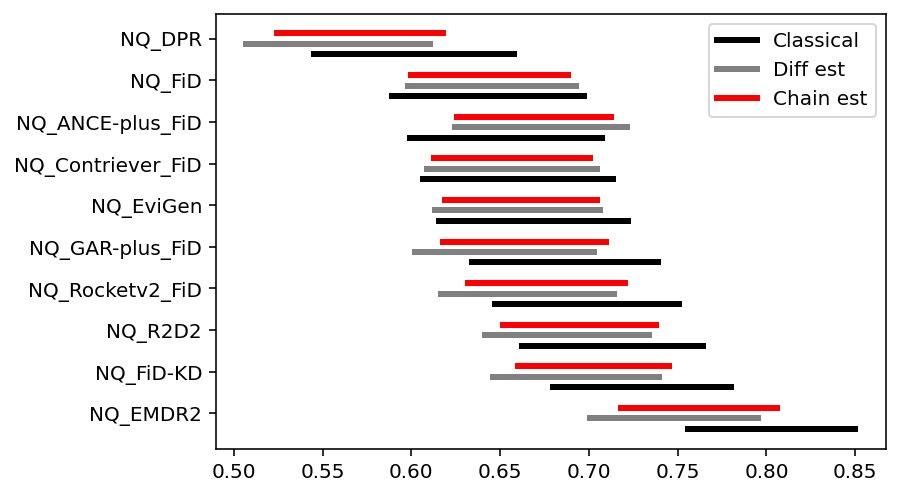}
\end{center}
\caption{Confidence intervals for the chain rule  estimate with abstentions, the difference estimate, and the classical method for the open-book QA methods from \cite{kamalloo-etal-2023-evaluating}.} \label{fig:kamaloo-cis}
\end{figure}

\begin{table}[t]
\centering
\begin{tabular}{lcl}
\toprule
            & mean interval & width ratio\\
            & width         & to classical\\
\midrule
classical   & 0.105 & 1.00 \\
difference estimate & 0.096 & 0.92\\
chain rule estimate & 0.089 & 0.85 \\
\bottomrule
\end{tabular}
\caption{Summary of confidence interval sizes at $n=300$ with an abstaining autorater for the data of \cite{kamalloo-etal-2023-evaluating}, averaged over 10 models.  Applying the difference estimate requires linearizing the autorater scores (see text).} \label{tab:kamalloo-summary-abstain}
\end{table}

\begin{table}[t]
\centering
\begin{tabular}{lcl}
\toprule
            & mean interval & width ratio\\
            & width         & to classical\\
\midrule
classical   & 0.105 & 1.00 \\
difference estimate & 0.095 & 0.91\\
chain rule estimate & 0.085 & 0.81\\
\midrule
 & \multicolumn{2}{c}{\textit{tuning on $S_n$}} \\
\midrule
difference est + ptune & 0.083 & 0.79 \\
\bottomrule
\end{tabular}
\caption{Summary of confidence interval sizes with a binary autorater for the data of \cite{kamalloo-etal-2023-evaluating}, averaged over 10 models.  The autorater was engineered to ``back off'' to exact match when the prompted model abstained (see text) } \label{tab:kamalloo-summary}
\end{table}

As summarized in Figure~\ref{fig:kamaloo-cis} and 
Table~\ref{tab:kamalloo-summary-abstain} confidence intervals produced by the chain rule estimate are about 15\% smaller than the classical intervals, and about 8\% smaller than the difference estimate intervals. \footnote{Notice that to use the difference estimate, it is necessary to map the unordered discrete values ``n'', ``y'', and ``u'' into real numbers: we made the natural assignment of $n=0$, $y=1$, and $u=0.5$, which makes the values similar to the binary 0/1 values used for the human rating.  Other options are discussed in Appendix~\ref{sec:linear}.}

For this dataset, there is another reasonable strategy for handling unparseable judge model responses, namely, ``backing off'' to the exact-match evaluation metric which is traditionally used for evaluation of open-book answers. Doing this improves both the chain rule estimate and the difference estimate slightly (as shown in the top Table~\ref{tab:kamalloo-summary}, but the chain rule still performs best.

If we consider tuned methods, we note that adding powertuning to the difference estimate improves performance significantly for these tasks. As shown in the bottom section of Table~\ref{tab:kamalloo-summary}, the power-tuned difference estimate slightly outperforms the chain rule estimate.

\subsubsection{Practical implications}

Although a 20\% reduction in width may seem modest, because confidence shrink with the square root of the number of labeled data, the number of labels needed may be reduced dramatically due to a small reduction in CI width.  Showing curves like those of Figure~\ref{fig:part} for all ten models is impractical, so
following \cite{doi:10.1126/science.adi6000} we subsampled the available human-labeled data to find the smallest sample size $n_\textit{min}$ that would give a confidence interval the same width or smaller than the baseline interval obtained by running the classical method on all available data.  The results, shown in Table~\ref{tab:kamalloo-min-n}, indicate that only 60\% of the data need be labeled to get estimates that are just as precise.  (These results use the best untuned model above, i.e., the chain rule estimate with backoff instead of abstentions.) 

We also estimated the number of examples $n^+$ that would need to be human-labeled to get the chain-rule estimate's interval width using the classical method.  The average $n^+$ for these tasks is 465, or about 57\% more labels (not shown in any table or figure).

\begin{table}
\centering
\begin{tabular}{llcc}
\toprule
{} &                 &  $n_\textit{min}$  with &  $n$ with \\
 &                    & chain rule & classical\\
 &                    & interval & interval \\
\midrule
 &             DPR &    151 &          291 \\
 &             FiD &    129 &          300 \\
 &   ANCE/FiD &    137 &          300 \\
 &    RocketQAv2/FiD &    185 &          299 \\
 &  Contriever/FiD &    141 &          300 \\
 &          FiD-KD &    150 &          300 \\
 &    GAR/FiD &    151 &          300 \\
 &          EviGen &    155 &          299 \\
 &           EMDR2 &    241 &          274 \\
 &            R2-D2 &    175 &          300 \\
 \midrule
 & average & 161.5 & 296.3 \\ 
\bottomrule
\end{tabular}
\caption{Minimal number of human-labeled examples $n_\textit{min}$ for the chain rule to obtain a confidence interval width smaller than the classical interval width using all available human-labeled examples. } \label{tab:kamalloo-min-n}
\end{table}

\subsubsection{Coverage tests} \label{sec:coverage}

Without large human-labeled sets, it is not possible to determine if the confidence intervals produced include the true values of $\theta^*$.  We thus used synthetic data 
to conduct coverage tests: for details, see Appendix~\ref{sec:kamaloo-coverage}.  Again we experimented with the best untuned model above. 

On 1000 synthetic datasets, the ``true'' $\theta^*_H$ used to generate the data was included in the 95\% confidence interval all but 48 times, or 95.2\% of the time. 




\subsection{Designing a PPI method for side-by-side tests} \label{sec:sxs-results}

One claimed advantage of the Bayesian approach is that it is straightforward to design a new PPI method.
We now consider constructed such a new method for side-by-side tests---to our knowledge, no PPI method for this task exists.

In side-by-side tests, raters (human $H$ or automatic $A$) label a pair $x$ with $w$, $l$, or $t$, for win, loss, or tie.  In the example of evaluating of question-answering systems, $x$ would contain a question and two answers (one from a baseline method, and one from a proposed improvement) and $f(x)$ predicts the human response with $\hat{y} \in \{w,l,t\}$. As before we assume a small set $S_n$ with both human and autorater labels and a larger dataset $\twid{S}_N$ with only autorater labels.

The goal of side-by-side tests is to see if the proposed method outperforms the baseline.  We formalize this with a statistic that tests to see if wins are more likely than losses:
\[ e(S_n, \twid{S}_N) \equiv \hat{\mu}_{Hw} - \hat{\mu}_{Hl}
\]
where
\begin{eqnarray*}
    \hat{\mu}_{Hw} & = & \prop{H=w}{S_n} \\
   \hat{\mu}_{Hl} & = & \prop{H=l}{S_n}   
\end{eqnarray*}
The designed statistic is a straightforward extension of the chain rule estimate.  We use the chain run to define autorater-assisted versions of $\hat{\mu}_{Hw}$ and $\hat{\mu}_{Hl}$ and then subtract them.
\begin{small}
\begin{eqnarray*}
   \hat{p}_{A=a} & = & \prop{A=a}{\twid{S}_N} ~,~ \mbox{for~} a\in\{w,l,t\} \\
   \hat{p}_{H=w|a} & = & \cprop{H=w}{A=a}{\twid{S}_N} ~,~ a\in\{w,l,t\} \\
   \hat{p}_{H=l|a} & = & \cprop{H=l}{A=a}{\twid{S}_N} ~,~  a\in\{w,l,t\} \\
 g(S_n,\twid{S}_N) & = &  \sum_{a \in \{w,l,t\}} \hat{p}_{H=w|a} \cdot \hat{p}_{A=a} ~ -  ~ \hat{p}_{H=l|a} \cdot \hat{p}_{A=a} 
\end{eqnarray*}
\end{small}

\subsubsection{Data and experimental procedure}

Freely available, high quality side-by-side labeled data for pairs of models is difficult to obtain.  However, when the same outputs of two different have been individually labeled, it is easy to simulate side-by-side ratings: e.g., if model $A$'s output on input $x$ is rated as ``good'' and model $B$'s output on $x$ is rated as ``bad'',  then one can rate $x$ as a ``win'' for $A$.  The same process can be used to simulate an autorater that produces side-by-side ratings.  Hence for for simulated side-by-tests, we used again the attributed QA data distributed by the authors of \cite{bohnet2023attributed}.

We began by finding a set of pairs of models $M_1, M_2$ such that (1) both models are competitive (2) neither model makes use of the autorater internally and (3) $M_1$ outperforms $M_2$.  We then used a multiple hypothesis test to find a set of 59 pairs $M_1, M_2$ where every $M_1$ outperforms $M_2$ with 95\% confidence (using a classical paired test on proportions and all the human data).  

We then reran the classical test and the method of Sec~\ref{sec:sxs} with subsamples of various sizes, ranging from 50 to 1000, as in the experiments above, and averaged over ten trials.

\subsubsection{Results and practical implications}

\begin{figure}[t]
\begin{center}
    \includegraphics[width=0.45\textwidth]{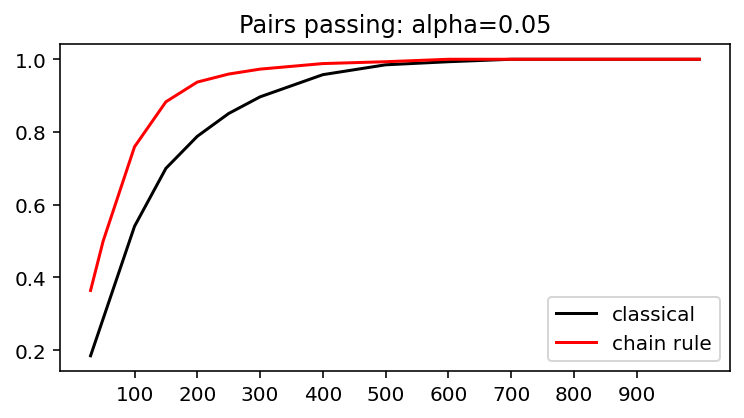}
\end{center}
\caption{Fraction of pairs of truly different systems that can be distinguished by a paired test, with classical methods and a chain rule estimate.} \label{fig:pairs}
\end{figure}

The results are shown in Figure~\ref{fig:pairs}.  The $x$ axis is the number of human-labeled examples used, and the $y$ axis is the fraction of the 59 pairs that can be separated at a 95\% confidence level.  We only compared the chain-rule test and a classical paired test, as previously published PPI tests are not directly applicable to paired tests,
and our prior experiments in linearizing discrete scores for difference estimates suggest that doing this effectively requires substantial exploration (see Appendix~\ref{tab:linearize}.)

As the figure shows, the chain rule test is dramatically different according to this measure: for example, with 100 examples, only 54\% of the pairs were separable with the classical test, verses 76\% with the chain rule test, and with 200 examples, the corresponding numbers are 79\% and 94\%.

\subsubsection{Coverage tests}  \label{sec:aqa-coverage}

We ran synthetic-data experiments similar to those of Section~\ref{sec:coverage} for the side-by-side estimates, as detailed in Appendix~\ref{sec:kamaloo-coverage}.  On 1000 synthetic datasets, the ``true'' $\theta^*_H$ used to generate the data was included in the 95\% confidence interval 94.5\% of the time, which is not statistically significant lower than the expected result ($p$-value of 0.46 using a classical test.)

\subsection{Non-Bayesian experimental results}

To clarify the claims of this paper, we do not suggest that the methods here improve on prior methods simply because they are Bayesian.  Rather, we claim that (1) for PPI, Bayesian approaches are analytically simpler, facilitating development of new PPI methods which are (2) experimentally preferable to prior PPI methods.

\subsubsection{Discussion}

Let us begin this section with a short discussion of Bayesian {\it vs.} frequentist approaches to confidence intervals.  Recall a Bayesian \emph{credible interval} for parameter $\theta$, for \emph{confidence level} $c$, is a pair of values $\ell$ and $u$ such that 
\[
    \int_{\ell}^u f(\theta) d\theta = c
\]
where $f(\theta)$ as the posterior of a random variable $\theta$ given data $D$, $f(\theta)=p(\theta|D)$.  In this framework, the interval $\ell, u$ for level $c$ means that $p(\ell \leq \theta \leq u) \geq c$, where the probability is taken over the possible values of the unknown parameter $\theta$.

In a classical (frequentist) confidence interval, the same relationship between $f$, $\ell$, and $u$, and $c$ holds, but $f(\theta)$ is not a posterior of anything---it's just a pdf, depending on the data $D$, that has been designed to satisfy the constraint above.  To be precise about the classical claim, let us use $\theta*$ for the unknown and fixed parameter value,
and write $f=f_S$ to show its dependence on $S$.  A frequentist construction for $f$ is correct if
\begin{equation} \label{eq:freq}
    \forall \theta*, \ell, u: 
 p_{S}(\ell \leq \theta* \leq  u) = 
   \int_\ell^u f_{S}(\theta) d\theta
\end{equation} 
where $p_S$ means that the probability is taken over samples drawn according to $\theta*$, i.e., $S \sim p(S|{\theta*})$.

To compare these two interpretations briefly, the Bayesian approaches require a posterior over $\theta$, which in turn requires a prior. This is a double-edged sword: you can inject subjective knowledge about $\theta$ easily if you have it, but you have to make up some sort of prior even when you don't.  However, credible intervals are easy to compute for many cases (as we see above).   In contrast, the classical interpretation requires (and allows) no prior on $\theta$, but is a bit harder to make sense of, and in our experience it is often confused with the Bayesian credible-interval interpretation (especially by ML researchers who have been exposed to Bayesian methods).  It's also technically difficult to establish that a method satisfies Eq~\ref{eq:freq}, although of course classical tests for most common situations exist.

We emphasize, however, that while there is definitely a \emph{conceptual} difference between treating the sample $S$ as a random variable (in the classical setting) and treating the parameter $\theta$ as a random variable (in the Bayesian setting), the practical differences are less dramatic.  Usually there is only one $\theta$ and one $S$, and unsurprisingly the computations for classical and Bayesian  work out to be quite similar: in each case, one picks a pdf $f(\theta)$ and snips off the tails.  For the case of means, the two pdfs are identical.  For the case of proportions, fitting a Beta with a Jeffrey's prior to a sample produces generally produces curves almost identical to the Gaussians used in the Wald method.\footnote{The exception is for small $n$ and extreme $\theta$, where an exact test would be preferred, and here the Beta closely approximates the binomial of Clopper-Pearson test.}

\subsubsection{Experimental results}

The discussion above suggests that the Bayesian credible intervals explored here might also perform well according to frequentist goals.  To evaluate this, we re-visited the coverage experiments of Section~\ref{sec:coverage}, where 1000 synthetic datasets were produced with different (but plausible) ``true'' parameters $\theta^*$.  As expected, about 5\% of these led to coverage failures---i.e., the chain-rule confidence interval did \emph{not} contain $\theta^*$.  In a Bayesian setting, this sort of failure is fine, since the probability for the interval is taken over different values of $\theta^*$; however, the frequentist interpretation of confidence intervals requires a 95\% interval contain the true value of the parameter for \emph{every} true value, where the probability is taken over samples---i.e., synthetically generated datasets.  

To evaluate performance according to the frequentist goal, we returned to the experiments of Section~\ref{sec:coverage},  randomly selected 20 of the $\theta*$'s which led to coverage failures in the initial set of 1000 trials, and re-generating 1000 synthetic datasets for each of these problematic $\theta^*$'s.
The fraction of coverage failures on this set of 20,000 simulations was 4.49\% overall.  Considering the 20 values for $\theta*$ separately, each of the 1000 $\theta*$-specific runs had between 37 and 54 coverage failures, showing the technique is well-calibrated in a frequentist sense as well.  We performed a similar\footnote{To reduce cost, here we ran a total of 1000 trials, where in each trial we selected one random coverage-failing $\theta^*$ from the 55 observed, and regenerated a dataset from that.} experiment with the data from Section~\ref{sec:aqa-coverage} and obtained similar results, with an overall coverage-failure rate of 3.9\%.

\subsubsection{The bootstrap as alternative to Monte Carlo integration}

Our approach also relies on Monte Carlo integration, a Bayesian approach.
An alternative to using Monte Carlo integration to evaluate CIs for proxy estimands is the bootstrap \cite{wasserman2013all}, which makes no explicit Bayesian assumptions.

The bootstrap can be used to compute expected values and confidence intervals over any numeric function $g(S)$ computed on a sample $S$.  In bootstrapping, one creates  $B$ \emph{bootstrap replicates}
$S^1, \ldots, S^B$ and then computes $g^1=g(S^1), \ldots, g^B=g(S^B)$.  A bootstrap replicate is simply a dataset of size $|S|$ that is formed by randomly sampling $|S|$ elements of $S$ with replacement.  The results of applying $g$ to each replicate are handled same way as the results of applying $g$ to samples from the posterior in Monte Carlo integration.  (In fact, our use of Monte Carlo integration is closely related to \emph{parametric bootstrapping} \cite{efron2012bayesian}.)

Compared to Monte Carlo integration, bootstrapping is slower, since samples of $g$ require time $O(N)$, instead of constant time.  However, bootstrapping does not require parameter independence, and does not require picking priors for each parameter.  Although the arguments behind the correctness of bootstrap are
different, it behaves similarly on these tasks.  For instance, we repeated the experiments of Table~\ref{tab:kamalloo-summary} using bootstrapping instead of Monte Carlo integration, and results that were numerically almost identical: e.g., the chain rule estimate's average width was measured as 0.084358 with the bootstrap, and 0.084516 with Monte Carlo integration.

\section{Conclusion}

One often-proposed approach to avoiding evaluation bottlenecks in developing LLM-based systems that produce long-form outputs is to use a secondary LLM-based system as a judge or ``autorater'' of input, output pairs.  Autoraters can be used to score outputs more cheaply than human raters, but are potentially biased.  To address this, \emph{prediction-powered inference (PPI)} methods \cite{doi:10.1126/science.adi6000} can be used, which combine a small number of human-rated outputs and a larger number of autorater-rated outputs, and produce a confidence interval that contains the average human rating, but is as small as possible.

Here we propose an new approach to PPI, where a target estimand (e.g., the mean human rating score) is coupled with a \emph{proxy estimate} that makes use of autoratings in a carefully designed, task-specific way.  We use a Bayesian formulation and simple general-purpose numerical methods to compute confidence intervals over the proxy estimand.  Using this approach leads to a number of novel proxy estimands that improve prior results on several tasks, sometimes dramatically, and which sometimes eliminate cumbersome engineering steps.  For instance, in evaluating two attributed QA models, the best existing PPI method reduces the CI width to 91\% and 86\% of the classical CI width, while our method reduces CI width to 78\% and 71\% of the classical width; if powertuning methods are allowed, prior methods obtain 83\% and 79\%, and our method obtains 76\% and 69\%.  On another set of tasks, prior PPI methods give 93\%, or 91\% with some engineering of the autorater, while
direct application of our method gives a reduction to 85\% of the classical CI width, implying that 60\% fewer labeled examples would be needed to obtain equally statistically meaningful results.  We also show a dramatic improvement in conducting side-by-side tests: e.g., with 200 examples, only 79\% of the truly different pairs of models were separable with the classical test, verses 94\% with the chain rule test.

\subsection{Limitations}

Our approach is Bayesian, not frequentist, and makes guarantees different from those associated with traditional confidence intervals.  Although the approach can also be implemented with non-Bayesian methods like the bootstrap, and that experimentally, the method seems to have good frequentist properties as well, this is still an important limitation, and makes the results difficult to compare directly with prior work in this area.  An important area for future work is developing frequentist versions of the successful methods explored here, notably the stratified PPI methods (and the closely-related chain rule estimates).

In prior work \cite{doi:10.1126/science.adi6000}
proposed PPI methods that apply to solutions to any convex optimization tasks, which are ideal for certain estimation statistics, such as parameters of linear or logistic regression, or Bradley-Terry coefficients \cite{boyeau2024autoeval}.  It is not immediately obvious if it is possible (or appropriate) to apply our approach to such problems.

\label{page:end}


\bibliography{diem}
\bibliographystyle{icml2022}

\newpage

\appendix
\section{Experimental Details}

\subsection{Python Implementation}  \label{sec:python}

\begin{table}
\begin{small}
\begin{verbatim}
import diem
def theta_fn(data):
  # pandas DataFrames holding S_n and S_N
  labeled_df, unlabeled_df = data
  return {
    'autorater_mean':
      diem.Mean(unlabeled_df.f),
    'rectifier':
      diem.Mean(labeled_df.y - labeled_df.f)
  }
def g_fn(autorater_mean, rectifier_mean):
  return autorater_mean + rectifier_mean
lo, hi = diem.MCIScorer(
  data, theta_fn, g_fn).score().ci()
\end{verbatim}
\end{small}
\caption{Sample Python code that implements the difference estimate.}  \label{tab:python}
\end{table}

Table~\ref{tab:python} illustrates an implementation of Monte Carlo integration in our software package, called \verb|diem| (for Design of Interval Estimation Methods). 
Implementing a PPI method is broken down into two steps: constructing and naming the $\theta$'s (means and proportions) that will be used in $g$, and then a Python implementation of the proxy estimand.  The \verb|Mean| object is 
responsible for sampling from the posterior, and the \verb|Scorer| object orchestrates sampling from each posterior, passing the samples to the proxy estimate function (here \verb|g_fn|), and creating an object to hold the results.

The code allows one to construct only three kinds of parameter posteriors, called \verb|Mean| (which has a Gaussian posterior for $n > 30$), \verb|Proportion| (a Beta posterior, using a Jeffrey's prior), and \verb|KProportion| (a Dirichlet posterior, for which we use the prior $\alpha_1=\ldots\alpha_K = \frac{1}{K}$.).
For \verb|Mean|s the Gaussian has a variance of $\hat{\sigma}^2/n$ where $\hat{\sigma}^2$ is the sample variance.  Instead of a Gaussian we use the appropriate Student's $T$ curve for $n<30$, which reflects standard classical statistical practice.

\subsection{Experiments: Seahorse and Attributed QA} \label{app:seahorse}

Regression trees were fit using the \verb|DecisionTreeRegressor| implementation from the \verb|sklearn.tree| package,
using the \verb|max_leaf_nodes| parameter to fix $K$.  
When tuning $K$, we considered the values
$2, 3, 5, 10, 20, 40$ and picked the one with smallest CI width.

The equal-frequency partitions were always selected using only the unlabeled data, and the tree is fit using only the labeled data.  This means that it is possible for there to be very few members of a partition in $S_n$ (for equal-frequency partitions) or $\twid{S}_N$ for tree partitions.  To fix this, we used a simple post-processing step, where (1) if there are any partitions that have fewer than 3 members in either $S_n$ or $\twid{S}_N$, those partitions are deleted, and replaced with a special ``miscellaneous'' partition and then (2) if the ``miscellaneous'' partition has fewer than 3 members in either part of the data, the smallest partition is deleted and its members are added to the ``miscellaneous'' one.

\begin{figure}[tb]
    \includegraphics[width=0.45\textwidth]{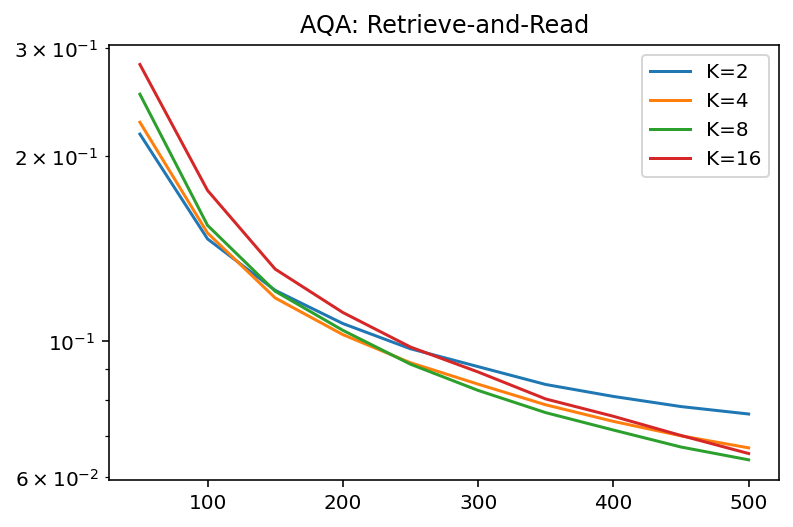}
\caption{Varying the number of partitions for one dataset.}
\label{fig:vary-part-k}
\end{figure}

In Figure~\ref{fig:vary-part-k} we vary the number of partitions for a single dataset (note the $y$-axis here is log scale, to make it easier to separate the results).  In general the best results are to use few partitions for small $n$, and more with larger $n$.

\subsection{Synthetic Data} \label{sec:kamaloo-coverage}

To generate synthetic data similar to the data of \cite{kamalloo-etal-2023-evaluating}
we used this procedure.
\begin{enumerate}
\item We created a Beta posterior distributions for $p(H|A)$, $p(H|\neg A)$, and $p(A)$ for each QA method at $n=300$.
\item To create a set of ``true'' values $\theta^*$ , we pick a QA method uniformly, and then sample values from the corresponding Betas.  This maintains dependencies that might exist in the parameter values, but also allows many different ``true'' $\theta^*$'s to be chosen.
\item Finally, we uniformly sample values of $n \sim \{100,\ldots,500\}$ and $N \sim \{3000,\ldots,4000\}$, and then generate a synthetic dataset of the right size using the ``true'' $\theta^*$ sampled in step 2.  
\end{enumerate}

We then run the chain rule estimate on these generated datasets and test for coverage.

For the experiments of Section~\ref{sec:aqa-coverage}, we used the same procedure, except instead starting with Beta distributions at $n=300$, we consider each $n$ used in creating the curves of Figure~\ref{fig:pairs} (i.e., $n=30, 50, 100, \ldots 300, 400, \ldots, 1000$) for a total of 826 different Betas.  To create true $\theta^*$ values, we uniformly sample $n \sim \{500, \ldots, 1000\}$ and $N \sim \{2000,4000\}$. 

\subsection{Linearizing the choices made by the abstaining autorater} \label{sec:linear}

\begin{table}
\centering
\begin{tabular}{lcl}
\toprule
            & mean interval & width ratio\\
            & width         & to classical\\
\midrule
~~chain rule estimate & 0.088 & 0.84 \\
~~difference estimate & 0.149 & 1.43\\
~~classical   & 0.104 & 1.00 \\
\midrule
\textit{$n=-1, y=+1, u=0$} &  & \\
~~difference estimate & 0.193 & 0.92 \\
~~classical   & 0.210 & 1.00 \\
\bottomrule
\end{tabular}
\caption{Alternative linearization schemes for abstaining autoraters.}
\label{tab:linearize}
\end{table}

The discrete values ``n'', ``y'', and ``u'' were internally encoded by the Dirichlet as integers 0, 1, and 2 respectively.  In addition to the scheme described in the paper, we considered several other linearization schemes, all of which were dramatically worse for the difference estimate, as summarized in Table~\ref{tab:linearize}.

We considered all six permutations of these codes and report numbers for the best in the top of the table.   It is a little surprising that the difference estimate is so much worse here worse than the classical estimate.  In our experiments we did note that the difference estimate does give an improvement over the classical method when you map $n$ to $-1$, $y$ to $+1$, and $u$ to $0$, but in this case all the CI widths are larger.



\end{document}